\documentclass[12pt, a4paper, twocolumn]{article}

\usepackage{geometry}
\title{Efficient Learning in Chinese Checkers: Comparing Parameter Sharing in Multi-Agent Reinforcement Learning}

\author{Allen Gu and Noah Adhikari}

\date{
    {\small University of California, Berkeley\\[2ex]} 
}

\usepackage{xurl}
\usepackage[super,comma,sort&compress]{natbib}
\usepackage{abstract}

\usepackage{lipsum}

\usepackage{hyperref}
\hypersetup{colorlinks=true, urlcolor=blue, linkcolor=blue, citecolor=blue}

\usepackage{array}
\usepackage{listings}
\usepackage[noline,ruled]{algorithm2e}
\usepackage{amsmath}
\usepackage{amsfonts}
\usepackage{graphicx}
\usepackage{multicol}

\usepackage{etoolbox}
\preto\align{\par\nobreak\small\noindent}
\expandafter\preto\csname align*\endcsname{\par\nobreak\small\noindent}

\begin{document}


\twocolumn[
  \begin{@twocolumnfalse}
    \maketitle
    
    \begin{abstract}
      We show that multi-agent reinforcement learning (MARL) with full parameter sharing outperforms independent and partially shared architectures in the competitive perfect-information homogenous game of Chinese Checkers. To run our experiments, we develop a new MARL environment: variable-size, six-player Chinese Checkers. This custom environment was developed in PettingZoo and supports all traditional rules of the game including chaining jumps. This is, to the best of our knowledge, the first implementation of Chinese Checkers that remains faithful to the true game.

      Chinese Checkers is difficult to learn due to its large branching factor and potentially infinite horizons. We borrow the concept of branching actions (submoves) from complex action spaces in other RL domains, where a submove may not end a player's turn immediately. This drastically reduces the dimensionality of the action space. Our observation space is inspired by AlphaGo with many binary game boards stacked in a 3D array to encode information.

      We use proximal policy optimization (PPO) to train all of our agents. Our experiments investigate three variations of parameter sharing over our agent's architectures: independent, shared-encoder, and fully-shared. The independent approach involves training six entirely separate agents, each with their own policy. The shared-encoder architecture involves sharing the encoder network between all agents, but each agent having its own policy and value function heads. Our fully-shared variation involves sharing all parameters between all agents. All policies use action masking to avoid illegal moves.

      Our experiments first analyze the performance of these three approaches on our Chinese Checkers environment for board size $N = 2$. We train policies through each of the three variations and evaluate them throughout the progression by simulating games against random opponents. Each random opponent samples a legal move uniformly at random. During evaluation, we introduce two metrics: win rate and game length measured in the number of turns to win.

      We find that all three variations achieve a 100\% win rate against the random opponents. The policy trained through full-parameter sharing achieves these results most efficiently, within the first 50,000 environment steps. The shared encoder model achieves a 100\% win rate soon after while the independent approach lags far behind. Additionally, we look at the metric of game length--how many turns a policy takes to win. In this metric, as well, full parameter sharing trains more efficiently and converges with better performance.
      
      For the trained policies, we briefly analyze game strategy and raise concerns over distributional shift in self-play, especially when using parameter sharing. Although our trained policy was still performant against random opponents, many possible states are unexplored through our training methods. Attempts to increase exploration through changing PPO's entropy coefficient led to inconclusive results and did not noticeably improve performance. We attribute the robustness of our trained policies to the fact that they fixate on their own pieces rather than the global board state.

      In the head-to-head-matches, we find that full parameter sharing considerably outperforms the partially-shared and independent architectures at all stages of training.
      
      We attribute the increase in performance to the fact that the fully shared model is able to take advantage of the homogenous environment to effectively learn six times as quickly compared to the independent approach. We do not expect the same performance gains in heterogenous environments, but similar methods in stochastic environments may prove effective.

      The PettingZoo environment, training and evaluation logic, and analysis scripts can be found on \href{https://github.com/noahadhikari/pettingzoo-chinese-checkers}{Github}.
    \end{abstract}
    
    \vfill
  \end{@twocolumnfalse}
]

\clearpage
\newpage
\section{Introduction}

Multi-agent reinforcement learning (MARL) has been a growing topic of interest in recent years \cite{GrowingMARLInterest}. Most prior work in this area has focused on two-player competitive, zero-sum games such as Go \cite{AlphaGo} or positive-sum cooperative games\cite{PositiveSumCoopGames}. Several frameworks and tools have been built to support these environments \cite{PettingZoo} \cite{Gym}. 

The behavior of agents in competitive environments with more than two players is much less well studied \cite{GrowingMARLInterest}. Stochastic games such as poker are popular in this domain, but complete-information environments are not as common, and it is difficult to find out-of-the-box environments for competitive multiplayer complete-information environments. In this paper, we develop a custom PettingZoo environment for Chinese Checkers, a six-player competitive homogenous complete-information game, as well as study the performance of several MARL architectures on this environment.

Though reinforcement learning for Chinese Checkers has been explored before \cite{ChineseCheckersRL1} \cite{ChineseCheckersRL2}, our implementation is, to the best of our knowledge, the first to remain faithful to the true game.

\section{Background}

\subsection{Chinese Checkers Rules \cite{ChineseCheckersRules}}

\begin{figure}[h]
  \centering
    \includegraphics[width=0.5\textwidth]{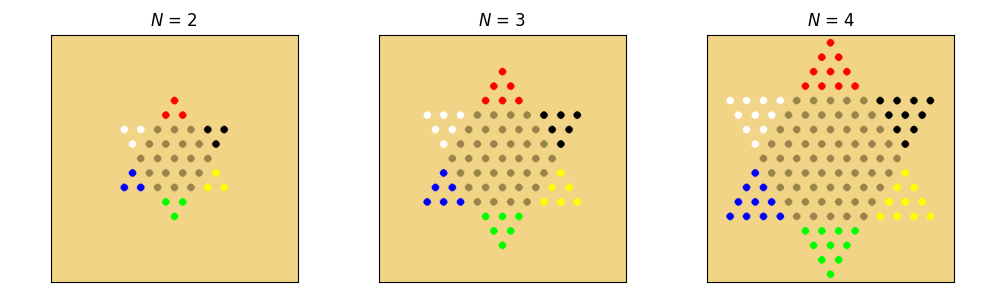}
  \caption{Game boards of size $N = 2, 3, 4$}
  \label{fig:boards}
\end{figure}

Chinese Checkers is a game traditionally played with 2, 3, 4, or 6 players of differing colors on a star-shaped board as seen in Figure \ref{fig:boards}. Each player traditionally has 10 pegs, which are initially placed in the corner of the board closest to them. The goal of the game is to move all of one's pegs to the corner of the board opposite to their starting corner.

\subsubsection{Movement}
The game is played in turns. On a player's turn, they must move one of their pegs to an adjacent empty space, or jump over an adjacent peg of any color to an empty space on the other side. A player may jump over multiple pegs in a single turn. A player can only move their own pegs. Only if a player has no possible moves may they pass their turn entirely.

\subsubsection{Winning}
The first player to move all of their pegs to the opposite corner wins the game. Though it is possible to continue playing after this point for ranking purposes, we focus on the variant where the game is terminated after the first player wins.

\subsubsection{Spoiling}
It is possible for a player to prevent another from winning by continously occupying one of the pegs in their goal corner, in a process known as ``spoiling.''. There are no official rules preventing spoiling \cite{ChineseCheckersRules}, and rules prohibiting it tend to be quite complicated, so we permit it in our implementation.

\subsection{Hexagonal Coordinate Systems}

The geometry of Chinese Checkers lends itself nicely to a hexagonal grid, but it is nontrivial to represent hexagonal grids, and there are various ways to do so. \cite{HexagonalCoordinates}

One way involves exploiting a symmetry between cubes and hexagons, where each axis of the cube corresponds to the projection in 2D-space of one of the three axes of the hexagon. One can then describe the position of a hexagon in terms of three coordinates $(q, r, s)$. Since the hexagonal grid lies in a plane, however, the coordinates are subject to the constraint $q + r + s = 0$. This is known as the cube coordinate system.

Because of the planar constraint, only two of the three $qrs$-coordinates are necessary. $(q, r)$ is a common choice; this is known as the axial coordinate system.

\subsection{Proximal Policy Optimization (PPO)}

Proximal Policy Optimization\cite{PPO} (PPO) is an on-policy method with an architecture similar to actor-critic algorithms. It addresses stability concerns in traditional actor-critic algorithms by limiting large policy updates. The algorithm accomplishes this through a clipped surrogate objective:

  $$
  \mbox{\footnotesize
  $L = \mathbb{E}\left[\min\left(\frac{\pi(a_t | s_t)}{\pi_{\text{old}}(a_t | s_t)} \hat{A_t}, \text{ clip}\left(\frac{\pi(a_t | s_t)}{\pi_{\text{old}}(a_t | s_t)},\, 1 - \epsilon,\, 1 + \epsilon \right) \hat{A_t}\right)\right]$
  }
  $$

In the objective, the advantage term incentivizes taking actions that are better than average. The ratio $\frac{\pi(a_t | s_t)}{\pi_{\text{old}}(a_t | s_t)}$ is a measure of how far the new policy deviates from the previous one. By clipping and minimizing with the non-clipped ratio, the objective also deters making large changes to the policy. 

\section{Methods}

\subsection{Chinese Checkers Environment}

We developed a custom PettingZoo environment for Chinese Checkers to support MARL. The board has the same shape as the traditional game, but can be configured to play with different sizes $N$, where $N$ is the side length of one of the corners of the board. For example, the traditional game has $N = 4$ and the $N = 2$ variation that we used for many experiments has 3 pegs per player.

\subsubsection{Number of Players}

For simplicity, we focus on the 6-player variant of Chinese Checkers as we wanted to investigate several algorithms in multiplayer self-play. However, our methods can be applied to other numbers of players.

\subsubsection{Turn Limits}

Unlike games such as chess, which impose turn limits to prevent infinite horizons \cite{ChessTurnLimit}, Chinese Checkers has no such rules. We impose a turn limit that ends the game with no winners after a large, configurable amount of turns (we found $\approx 200$ to be a good choice for $N=2$ and $\approx 1000$ to be a good choice for $N=4$).

\subsubsection{Submoves}

In Chinese Checkers, it may be difficult to enumerate all of an agent's possible moves, as jumps increase the number of possible moves significantly. Additionally, within a single move, it is possible to infinitely stall the game by jumping in a cycle.

To prevent this, we adopt the concept of branching actions from complex action spaces in other RL environments \cite{BranchingActions}, where a move consists of one or more submoves that may not end the player's turn immediately. In Chinese Checkers, a submove is a single jump, adjacent-space movement, or end-turn action. Jump submoves may not necessarily end a player's turn, but to prevent infinite turns, players cannot jump back to a space they previously occupied on that turn. Players can end their turn following a jump using the end-turn submove.

\subsubsection{Board Representation}
The game board uses cube coordinates, with the center of the board being $(q, r, s) = (0, 0, 0)$.

Note that due to the star-shaped nature of the board, it is difficult to represent the board as a perfect hexagon. To circumvent this, we simply treat the board as a hexagon of diagonal length $(4N+1)$, which corresponds to a circumscibed hexagon about the star. Extraneous spaces are then masked out to prevent movement out-of-bounds. 


For symmetry, we rotate the board so that the current player's home triangle is at the top of the board when making moves. 
  
The 3D representation of the board is quite sparse, so we flatten the board with the axial coordinate constraint $q + r + s = 0$ when communicating with agents.

\subsubsection{Observation Space}

The observation space is a flattened representation of 8 different layers of the board, where each layer provides unique information about the game state. Each layer is as a two-dimensional $(4N + 1, 4N + 1)$ grid where row and column indices $(i, j)$ map to axial coordinates $(q, r)$ of the board space. To maintain rotational symmetry, the board is always rotated when creating the observation so that the current player's home triangle is at the top of the board. The layers represent the following binary information for each space on the board:

\begin{center}
\begin{tabular}{|m{0.05\textwidth}|m{0.35\textwidth}|}
  \hline
  Layer & Description \\
  \hline
  0 & Current player's pegs \\
  1-5 & Other players' pegs, in clockwise order starting from the current player \\
  6 & Where the current player has jumped from this turn \\
  7 & Where the current player jumped during the last submove, if the last submove was a jump \\
  \hline
\end{tabular}
\end{center}

After flattening, the observation space becomes a one-dimensional $(4N + 1) \times (4N + 1) \times 8$ binary vector.


\subsubsection{Action Space}

Actions in the Chinese Checkers environment are represented by a tuple $(q, r, \texttt{direction}, \texttt{is\_jump})$:

\begin{center}
  \begin{tabular}{|m{0.1\textwidth}|m{0.3\textwidth}|}
    \hline
    $(q, r)$ & Position of the peg being moved in axial coordinates \\
    \hline
    \texttt{direction} & Direction of movement from $0$ to $6$ in angle increments of $\frac{\pi}{3}$, where $0$ is due-right, $1$ is up-right, etc. \\
    \hline
    \texttt{is\_jump} & Whether the action is a jump \\
    \hline
  \end{tabular}
  \end{center}

  If a previous submove was a jump or if no legal moves are available, a player may also end their turn immediately without a move (known as the end-turn action). Thus, we represent the action space as a discrete space of size $(4N + 1) \times (4N + 1) \times 6 \times 2 + 1$.

\subsubsection{Action Masking}

For each agent, the observation, in addition to the observation space, also includes an attribute \texttt{action\_mask} which is a binary vector of size $(4N + 1) \times (4N + 1) \times 6 \times 2 + 1$ that indicates which actions are legal for the current agent during the current submove. This is used to mask out illegal actions when sampling from the action space.

\subsubsection{Rewards}

Various reward schemes were tried:

\begin{enumerate}
  \item \textit{Sparse:} $+5N$ for winning, $-N$ for losing, $0$ otherwise
  \item Sparse reward, plus a small \textit{goal bonus} for entering pegs in the goal ($+0.1$) and penalty for removing them from the goal ($-0.1$)
  \item Sparse reward, plus a tiny \textit{movement bonus} for moving forward ($+0.001$) and a penalty for moving backward ($-0.001$)
  \item \textit{Positive-sum:} $+5N$ for winning, $0$ otherwise, plus goal and movement bonuses and penalties
\end{enumerate}

We found that agents with sparse rewards had difficulty learning, so we ultimately opted for a positive sum, dense reward scheme (4) due to our goals involving self-play.

\subsubsection{Environment Stepping}

The pseudocode for a single environment step within our game is summarized below. Each step corresponds to moving a single peg within the game or ending a turn. Consecutive jumps are handled across multiple environment steps.

\begin{algorithm}
  \SetKwInOut{Input}{Input}

  \textbf{function} step (a)
  
  \Indp

  \Input{\textit{a}: Integer action in $[0, (4N + 1) \times (4N + 1) \times 6 \times 2 + 1)$}
  
  Convert integer action to a submove

  Rotate the board to the current player and perform the submove

  Update rewards for agents

  \If{submove is not a jump}
    {
      Advance to the next player
    }

  \Indm

  \textbf{end}

  \caption{Environment stepping in Chinese Checkers}
\end{algorithm}
  
\subsection{Multi-Agent Learning}

There are many approaches to configuring models in multi-agent settings. In this section, we describe both our model algorithm and the variations we tested on the Chinese Checkers environment.

\subsubsection{Model Architecture}

All agents learn through Proximal Policy Optimization (PPO) with a clipped surrogate objective\cite{PPO}. In our architecture, the actor and critic share an MLP encoder but have differing policy and value function heads.

\textbf{Encoder Network:} The encoder network takes in a vector the size of the observation space as input. Each encoder contains a hidden layer of size 64 and output layer of size 64, with ReLU activation functions.

\textbf{Policy Head:} The policy head is a single fully connected network of shape $(64, (4N + 1) \times (4N + 1) \times 6 \times 2 + 1)$ that takes the encoder output as input and outputs action logits.

\textbf{Value Function Head:} The value function head is a single fully connected network of shape $(64, 1)$ that takes the encoder output as input and outputs a scalar estimating the value of the state.

\subsubsection{Variations on Multi-Agent Setup}

We compare three different policy configurations for the multiple agents: fully-independent, shared-encoder, and fully-shared variations.
\begin{itemize}
  \item \textit{Fully-Independent:} Agents learn separately with their own single-agent PPO policy. 
  \item \textit{Shared-Encoder:} All agents share the same encoder, but have unique policy and value function heads.
  \item \textit{Fully-Shared:} Agents share all parameters, effectively training a single policy.
\end{itemize}

\subsubsection{Training Procedure}

We trained all three policy configurations using self-play. As the policies are updated, the agents become stronger and thus have stronger opposition, so this algorithm is reminiscient of curriculum learning \cite{SelfPlayCurriculumLearning}. The training algorithm is summarized below:



\begin{algorithm}
  \SetKwInOut{Input}{Input}
  \SetKwFor{RepTimes}{repeat}{times}{end}

  \textbf{function} train (n, policy)

  \Indp
  
  \Input{\textit{n}: Number of training iterations \\
         \textit{policy}: Policy configuration}
  
  Initialize 6 policies following the given configuration, sharing parameters if necessary

  \RepTimes{n} {
    Sample a batch of experiences by playing against the other policies

    Update the models using PPO
  }

  \Indm
  \textbf{end}
  \caption{Agent self-play training procedure}
\end{algorithm}

\subsubsection{Evaluation Procedures}

We evaluate the performance of trained agents via several metrics.

\textit{Random agent evaluation}: The evaluated agent plays against a field of random agents. The number of wins and the number of turns taken to win are both recorded.

\begin{algorithm}
  \SetKwInOut{Input}{Input}
  \SetKwInOut{Output}{Output}
  \SetKwFor{RepTimes}{repeat}{times}{end}

  \textbf{function} evaluate (n, policies)

  \Indp
  
  \Input{\textit{n}: Number of games \\
         \textit{policies}: Six configurations}
  \Output{\textit{wins}: Wins per policy \\
          \textit{lengths}: Game lengths}

  \RepTimes{n} {
    Play a game with the given policies

    Record whether the game was a win for each policy into \textit{wins}

    Record the number of turns taken for the game to terminate into \textit{lengths}
  }

  \Return{wins, lengths}

  \Indm

  \textbf{end}

  \caption{Evaluation procedure}
\end{algorithm}

\textit{Head-to-head matches}: Each of the three architectures contributes two agents in a match of six players. The positions of the agents are randomized, and the agents play several games against each other. The number of wins per player and the number of turns taken to win are both recorded.

\section{Experiments}

\subsection{Comparison of Multi-Agent Configurations}

\begin{figure*}[t]
  \centering
    \includegraphics[width=0.9\textwidth]{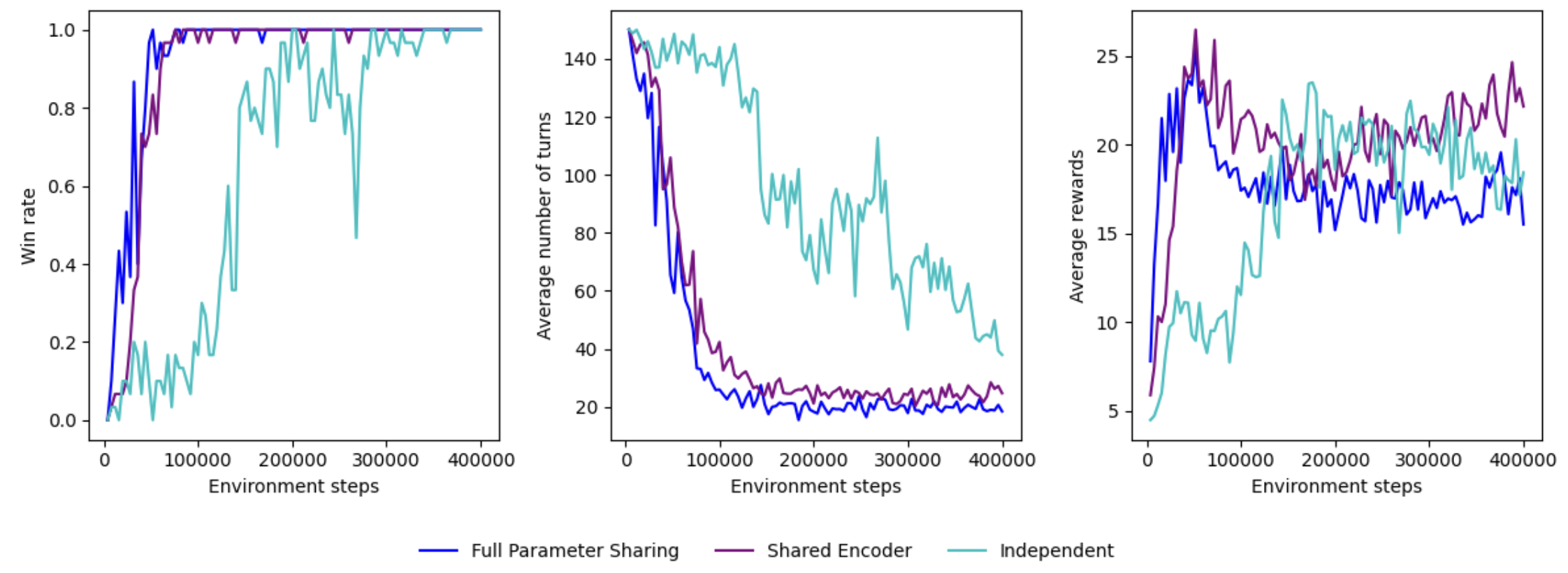}
  \caption{Evaluation occurred at repeated intervals throughout training for fully-independent, shared-encoder, and fully-shared multi-agent configurations. \textbf{Left:} Win rate of policy against five random policies. \textbf{Middle:} Average game length (turns made by policy). \textbf{Right:} Average rewards.}\label{fig:win_rate_game_length}
\end{figure*}

First, we compare the different multi-agent setups: fully-independent, shared-encoder, and fully-shared. Each algorithm was trained over 100 iterations of 4000 timesteps each, with evaluation after every timestep. During evaluation, one trained policy would compete for 30 rounds against five random policies until one player wins or 150 turns have been played.

From the results in Figure \ref{fig:win_rate_game_length}, we find that as more parameters are shared between the agents, the more efficient our training becomes. Win rate against random opponents is a better metric of model performance compared to average rewards because although longer games are less optimal, they tend to receive larger rewards.

Full parameter sharing reached over a 90\% winrate in less than 50,000 environment steps of training while the fully-independent algorithm took almost 200,000 environment steps. Note that the winrate of an algorithm excludes games that were truncated due to the turn limit.

Additionally, the game durations during evaluation were shorter for parameter-sharing methods. The policy trained through a fully-shared approach was able to win in fewer turns, where one turn consists of all moves made by a single player before allowing the next player to act. From Figure \ref{fig:win_rate_game_length}, full parameter sharing converges to a policy that wins in nearly 20 moves, faster than the other two approaches.

A naive strategy for Chinese Checkers could be to move each piece from the home triangle to the target triangle one by one, without jumps. For a board of size $N = 2$, this strategy performs 20 moves. The optimal strategy including jumps would improve this by a couple moves. Interestingly, even when facing opponents that may both harm (e.g. block the target triangle) or help (e.g. increase the potential for jumps), our model is near this hypothetical optimum.

\subsection{Analyzing Policy Strategies}

\begin{figure}[h]
  \centering
    \includegraphics[width=0.47\textwidth]{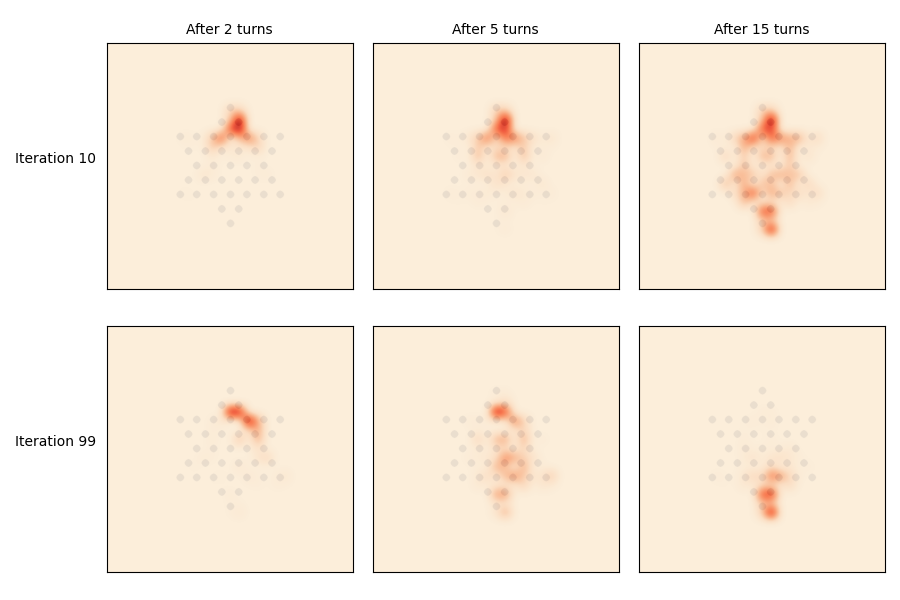}
  \caption{Heatmaps for the policy trained through full parameter sharing. These plots show the frequency of different peg locations on the board after some amount of turns have been made. The trained policy plays as red with pegs starting at the top triangle.}
  \label{fig:heatmap}
\end{figure}

Although all trained models eventually reach a 100\% win-rate against random players, policies tended to move the pieces to the target zone in an independent fashion: moving one piece first, then the next, and so on. To visualize these patterns, we run training checkpoints 100 times against five random policies. During each game, the board states were collected at different numbers of turns. These board states were used to generate heat maps in Figure \ref{fig:heatmap} showing the frequency of pegs throughout the board.

This behavior seems to be learned early on in training, as shown by Figure \ref{fig:heatmap}. By the 10th iteration, turn 15 shows a high probability that some pegs are still in the home triangle while others have already made it to the target triangle. A similar pattern can still be seen in the 99th training iteration, though at an earlier number of turns. In turn 5, many pegs are still left in the home triangle while others are far down the board.

\subsection{Head-to-Head-to-Head Matches}

\begin{figure}[h]
  \centering
    \includegraphics[width=0.47\textwidth]{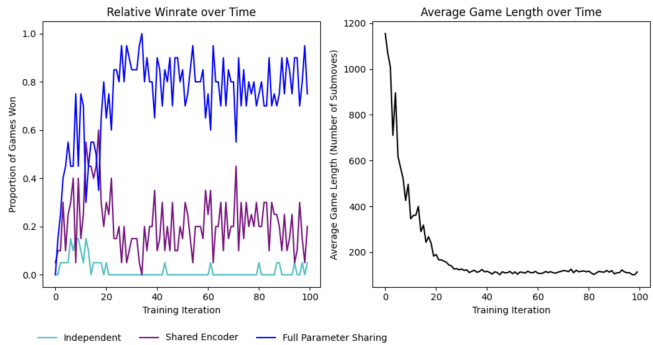}
  \caption{Win rates and game length of all three architectures piloting two randomized players per game throughout training.}\label{fig:archmatches}
\end{figure}

For head-to-head matches, we trained each of the three architectures for 100 iterations of 4000 timesteps each. After each iteration, we evaluated the performance of each policy by having it play 20 games against the other two policy configurations, with a turn limit of 200. Note that truncated games with no winner are not treated as wins for any player, hence the winrates may not sum to 1. The results are shown in Figure \ref{fig:archmatches}.

The fully-shared policy outperforms the other two policies when they are pitted against one another. Over time, the policies also complete games more quickly, indicating more efficient strategies.

We hypothesize that the fully-shared policy is able to take advantage of the homogenous environment to effectively learn six times as quickly compared to the independent approach. We expect full parameter sharing to still perform well in stochastic homogenous environments, but heterogeneous environments will probably not see the same performance gains.

\subsection{Exploration}

Distributional shift between training and evaluation is a concern for multi-agent self play. The distribution of observed states in self-play is not representative of all potential opponent policies. In the context of Chinese checkers, as a self-play agent improves, the board state will tend to follow certain patterns dependent on its learned policy. As a result, out-of-distribution states (such as those from a random policy) may harm agent performance during evaluation.

We hypothesized that encouraging exploration when parameter-shared MARL algorithms are trained in self-play would improve performance in the random agent evaluation. In PPO algorithms, exploration can be encouraged by an entropy coefficient $c$. This entropy term is added to the clipped surrogate objective. However, testing this claim through a hyperparameter sweep on $c$ yielded inconclusive results.

$$
  \mbox{\footnotesize
  $L = \mathbb{E}[L^{CLIP} + cS(\pi_{\theta}(s_t))]$
  }
$$

\begin{figure}[h]
\centering
    \includegraphics[width=0.4\textwidth]{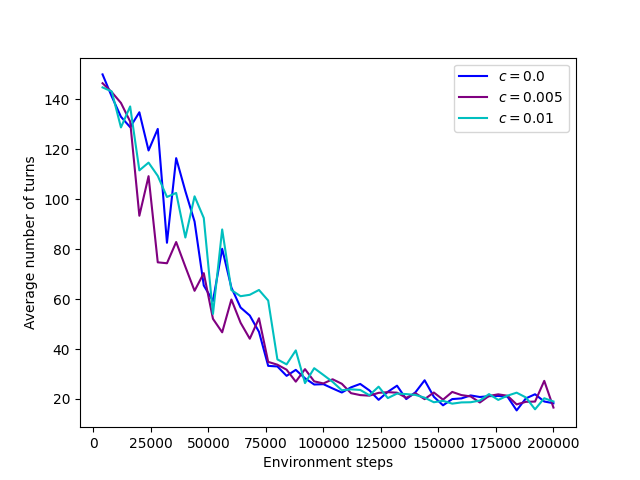}
  \caption{Effect of varying the entropy coefficient $c$ on average game lengths against random opponents.}\label{fig:exploration}
\end{figure}

PPO agents with full parameter sharing for $c = 0.0, 0.005, 0.01$ were trained for 50 training iterations of 4000 environment steps each and evaluated against random opponents. The average evaluation game length curves for the different hyperparameter values are shown in Figure \ref{fig:exploration}.

Due to the variance in game lengths, it was difficult to draw meaningful conclusions about the effect of exploration on parameter sharing approaches to MARL. We posit that the agents may disproportionately fixate on their own pieces rather than the global board state in trying to get to the goal so they are more robust to distributional shift in evaluation.

\subsection{Extension to Larger Boards}

\begin{figure}[h]
  \centering
    \includegraphics[width=0.47\textwidth]{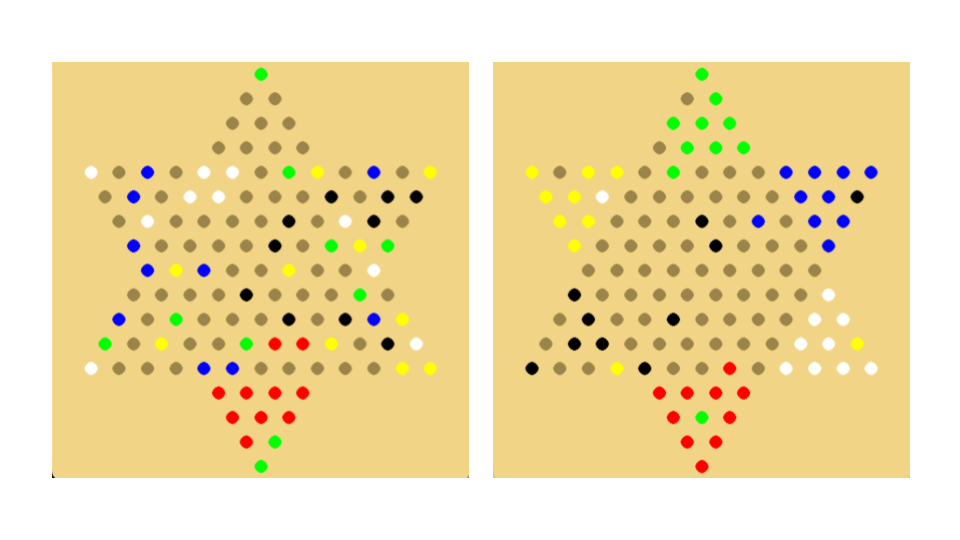}
  \caption{End game states for boards of size $N = 4$. A full parameter sharing policy trained for 100 iterations plays as red against (\textbf{left}) random opponents and (\textbf{right}) itself.}\label{fig:largerboards}
\end{figure}

For larger boards $N > 2$, the spoiling mechanic in Chinese Checkers becomes more prominent which prevented good results to compare the different multi-agent schemes. Although we tried experiments on larger boards, pegs would often become stuck or spoil other agents, rendering metrics like win rate less useful because most games end without a winner. Figure \ref{fig:largerboards} shows end game states for $N = 4$.

\section{Conclusion}

We have introduced a new competitive environment for multi-agent reinforcement learning. On this environment, we showed that even in multi-agent competitive settings, parameter sharing over multiple agents can improve training efficiency and performance. We found that PPO agents trained with full parameter sharing performed the best when using game length as an evaluation metric. A shared encoder training approach also performed better than the fully independent setup. We attribute these performance gains to the homogenous environment, for which parameter sharing is particularly well-suited.

\subsection{Group Contributions}

We collaboratively developed the PettingZoo Chinese checkers environment and MARL training and evaluation logic during synchronous work sessions. We also worked together on all sections of the report. \textbf{Noah Adhikari} extended the evaluation framework to compare the trained agents in head-to-head gameplay. \textbf{Allen Gu} analyzed the performance of the independent, shared-encoder, and fully-shared methods in play against random opponents and also coded the heatmap plotting to analyze the policies.


\nocite{*}
\bibliographystyle{ieeetr}
\bibliography{main}

\end{document}